# Automating Surgical Peg Transfer: Calibration with Deep Learning Can Exceed Speed, Accuracy, and Consistency of Humans

Minho Hwang, Jeff Ichnowski, *Member, IEEE,* Brijen Thananjeyan, *Member, IEEE,*
Daniel Seita, *Member, IEEE,* Samuel Paradis, Danyal Fer,
Thomas Low, *Member, IEEE,* and Ken Goldberg, *Fellow, IEEE*

*Abstract*—Peg transfer is a well-known surgical training task in the Fundamentals of Laparoscopic Surgery (FLS). While human surgeons teleoperate robots such as the da Vinci to perform this task with high speed and accuracy, it is challenging to automate. This paper presents a novel system and control method using a da Vinci Research Kit (dVRK) surgical robot and a Zivid depth sensor, and a human subjects study comparing performance on three variants of the peg-transfer task: unilateral, bilateral without handovers, and bilateral with handovers. The system combines 3D printing, depth sensing, and deep learning for calibration with a new analytic inverse kinematics model and time-minimized motion controller. In a controlled study of 3384 peg transfer trials performed by the system, an expert surgical resident, and 9 volunteers, results suggest that the system achieves accuracy on par with the experienced surgical resident and is significantly faster and more consistent than the surgical resident and volunteers. The system also exhibits the highest consistency and lowest collision rate. To our knowledge, this is the first autonomous system to achieve "superhuman" performance on a standardized surgical task. All data is available at
https://sites.google.com/view/surgicalpegtransfer.

*Note to Practitioners*—This paper presents a new approach to calibrating cable-driven robots based on a combination of 3D printing, depth sensing, inverse kinematics, convex optimization, and deep learning. The approach is applied to calibrating the da Vinci, commercial surgical-assist robot, to automate a standard "pick and place" task. Experiments suggest that the resulting system matches human surgical expert performance in speed and accuracy and significantly outperforms humans in terms of consistency. All details on the system including CAD models, code, and user study data are available online.

## I. INTRODUCTION

It is widely recognized that minimally invasive surgery can reduce post-operative patient pain and length of stay in the hospital [1]. Robotic Surgical Assistants (RSAs) such as the da Vinci improve ergonomics so that surgeons can perform minimally invasive surgery with improved dexterity and visualization through local teleoperation [2]. RSAs also have potential to actively assist surgeons using supervised autonomy of specific subtasks under close supervision to reduce surgeon fatigue and tedium [3], [4]. RSAs are cable-driven to fit within narrow abdominal portals, introducing hysteresis and tracking errors that make it challenging to automate surgical subtasks [5-8]. These errors are difficult to detect and track because encoders are located far from the joints. We present a novel approach to automating a high-precision surgical subtask by leveraging recent advances in depth sensing, recurrent dynamics modeling, and trajectory optimization.

We consider peg transfer, a standardized task from the Fundamentals of Laparoscopic (FLS) [9] surgeon training suite. The task setup is shown in Fig. 1. This paper makes 5 contributions:

1) The deep recurrent neural network (RNN) models applied in Hwang et al. [10] require training data consisting of trajectories that are similar to those encountered in the target task, so using a different trajectory distribution for training results in a performance drop. In this paper, we densely sample states on randomly-generated trajectories for training, and demonstrate that this achieves comparable performance on peg transfer without requiring task-specific trajectories, by interpolating the task trajectory of the robot at the same sampling interval as the trajectory for training.

2) While Hwang et al. [10] and Paradis et al. [11] are able to meet human-level success rates on the peg transfer task, these approaches are 1.5x to 2x slower than a skilled human operator,

Manuscript received April 13, 2022;
Minho Hwang is with the Department of Robotics Engineering, Daegu Gyeongbuk Institute of Science and Technology (DGIST), Daegu, Republic of Korea (e-mail: minho@dgist.ac.kr).
Jeff Ichnowski, Member, Brijen Thananjeyan, Daniel Seita, Samuel Paradis, and Ken Goldberg are with the Department of Electrical Engineering and Computer Sciences, University of California, Berkeley, USA
(email: jeffi@berkeley.edu, bthananjeyan@berkeley.edu, seita@cs.berkeley.edu, samparadis@berkeley.edu, goldberg@berkeley.edu)
Danyal Fer is with the Department of General Surgery, University of California San Francisco East Bay, Oakland, CA 94602, USA
(Danyal.Fer@ucsf.edu)
Thomas Low is with the Robotics Laboratory, SRI International, Menlo Park, CA, 94025, USA (thomas.low@sri.com)

respectively. In this paper, we optimize the robot arm trajectories using convex optimization. Furthermore, while human operators may have trouble executing multiple tasks simultaneously, a robot can parallelize task computation and execution using multiple concurrent processes. In contrast, prior work [12] considers only a basic bilateral setup without coordination between the arms, resulting in slower performance.

3) A novel closed-form solution that allows for fast inverse kinematics (IK) calculation. General IK solvers typically treat this problem as an iterative optimization problem and thus may take many iterations to converge to a solution. We observe that a closed-form analytic solution exists due to the special kinematic constraints of the surgical robot. This method can be applied to any 6 DoF surgical manipulator with Remote Center of Motion (RCM) mechanism.

4) The full FLS peg-transfer task that human surgeons train on requires the blocks to be handed over from one arm to the other before placing them onto the pegs. This paper also considers handovers, the full FLS peg-transfer task.

5) A detailed user study suggesting that the fully autonomous system achieves significantly fewer collisions and greater consistency than humans.

## II. RELATED STUDIES

### A. State Estimation

Trained human surgeon controls RSAs via teleoperation and compensates for cable-related effects by observing and reacting to robot motion. Surgical tasks often require positional accuracy of the end-effector in the workspace within 2 mm, and this is difficult to autonomously obtain with cable-driven surgical arms due to effects such as cable tension, cable stretch, and hysteresis [13], and such effects are exacerbated with flexible arms [14] and usage-related wear. To compensate for these errors, prior methods use techniques such as unscented Kalman filters to improve joint angle estimation [7] by estimating cable stretch and friction [6], or by learning fixed offsets for robot end-effector positions and orientations [5, 8, 15]. Peng et al. [16] proposed a data-driven calibration method for the Raven II surgery robot that uses three spheres and four RGB cameras to estimate the end-effector position.

In contrast to the previous works, we consider the problem of estimating the joint configuration, which can be incorporated more directly in collision checking, and we also learn to predict the commanded input given a history and desired joint angle.

### B. Surgical Task Automation

No surgical procedures in clinical settings today use autonomy, but researchers have studied automating surgical subtasks in laboratory settings. For example, prior researchers have shown promising results in automating peg transfer [12, 17-21], suturing [22-26], debridement [27, 28], clearing the surgical field [29], tumor ablation [30], and cutting gauze [31, 32]. There are attempts to automate surgical subtasks in in-vivo setting such as steering of endoscope [33, 34] and intestinal anastomosis with commercial robot manipulators and a specialized needle tool [35].

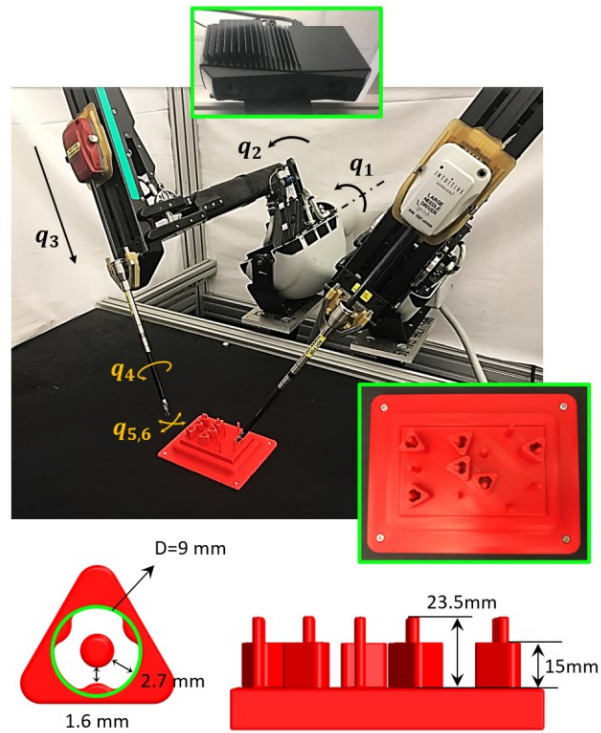

Fig. 1. Automated peg-transfer task: We use the da Vinci Research Kit (dVRK) robot from Intuitive Surgical with two arms. The blocks, pegs, and peg board are monochrome to simulate a surgical setting. The dimensions of the pegs and the blocks are shown in the lower left, along with a top-down visualization of the peg board to the lower right. The robot takes actions based on images taken from a Zivid depth camera, installed 0.5 m from the task space and 50 ° inclined from vertical.

### C. Surgical Peg Transfer Task

The FLS consists of 5 tasks designed to train and evaluate human surgeons performing minimally invasive surgery, such as peg transfer, precision cutting, ligating loop, and suturing with either extracorporeal or intracorporeal knots. In this paper, we focus on the first task, peg transfer, in which the goal is to transfer six triangular blocks from one half of a pegboard to the other, and then back (Fig. 2). For each block, this process requires grasping the block, lifting it off of a peg, handing it to the other arm, moving it over another peg, and then placing the block around the targeted peg. Since each block's opening has just a 4.5 mm radius, and blocks must be placed on cylindrical pegs which are 2.25 mm wide, the task requires high precision, making it a popular benchmark task for evaluating and training human surgeons [36-39].

Rosen and Ma [17] were the first to automate a version of peg transfer. Using one robot arm from the Raven II surgical robot [40] to transfer 3 blocks to the nearest pegs in one direction, they compared performance over 20 trials of the autonomous robot versus a human teleoperator. Their results suggested that the autonomous robot attained a 93.3 % block transfer success rate with an average of 8.3 seconds per transfer, as compared to the human with 100.0 % with an average of 16.3 seconds per transfer. In this work, we focus on the bilateral peg transfer task with 12 block transfers across both directions and handovers.

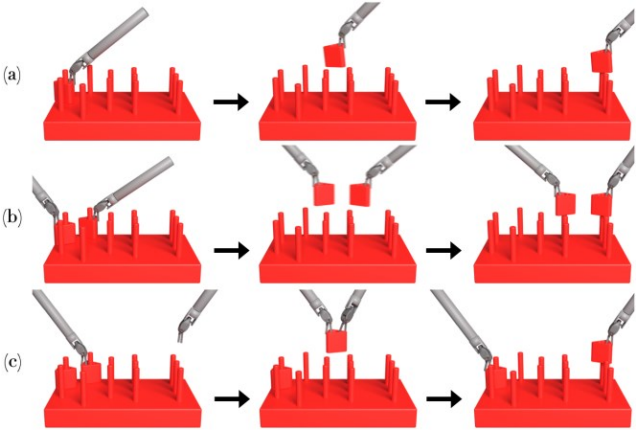

Fig. 2. Three variants of the FLS peg transfer task. We consider the three variants of peg transfer. The pegs and blocks are 3D-printed in red ABS at the same scale as the commercial FLS training system to be consistent and similar in color to in-vivo conditions. (a) Unilateral: block transfers performed by a single da Vinci arm. (b) Parallel Bilateral: block transfers performed by two arms in parallel. (c) Bilateral Handover: block transfers performed by both arms, with a hand-off between arms during each transfer.

Hwang et al. [12] developed the first system for a surgical robot to autonomously perform a variant of the six block FLS peg-transfer task using depth sensing and manual checkerboard-calibration. The system moved the robot via open loop commands to pick and place blocks from one side of the peg board to another without handover of the blocks between the two arms. In subsequent work, Hwang et al. [10] improved the accuracy of the physical robot using deep dynamics models trained on robot trajectories collected using fiducial markers and depth sensing, and improved the perception pipeline for peg transfer.

Paradis et al. [11] proposed Intermittent Visual Servoing (IVS), a paradigm for using a coarse open-loop controller in free space, while using learned visual servoing when the robot is close to an area of interest, such as when picking or placing a block. Paradis et al. learned visual servoing from demonstrations, and in peg-transfer experiments, showed that the learned model can be robust to tool changes. However, IVS incurred additional time delays during the correction phase.

### III. PROBLEM DEFINITION

#### A. Task Definition

We use red 3D-printed blocks and a 3D-printed pegboard (see Fig. 1). The red setup simulates a surgical environment where blood is common and surgeons must rely on subtle visual cues to perceive the state. To perform the task, the robot moves each of the 6 blocks from a peg on the left to a peg on the right, then moves all 6 back again. Unlike prior work, we also consider bilateral variants of this task with handovers between two arms.

We define the peg-transfer task as consisting of a series of subtasks with associated success criteria:

1) **Pick**: the robot grasps a block and lifts it off a peg.
2) **Handover**: the robot passes a block from one arm to the other.
3) **Place**: the robot places a block onto a target peg.

We define a transfer as a successful pick followed by a successful place with a handover for Bilateral Handover. The 6 blocks are initially placed on the left 6 pegs of the pegboard. In a trial of the peg-transfer task, the robot attempts to move all 6 blocks to the right 6 pegs and then moves them back. A trial is considered successful if all 12 transfers are successful. A trial can have fewer than 12 transfers if failures occur during the first 6 transfers and a block is irrecoverable, which is not considered as a successful trial.

We define the following variants of the peg-transfer task, illustrated in Fig. 2:

1) **Unilateral**: all 12 transfers are performed by a single arm. This is the variant considered in most prior work.
2) **Parallel Bilateral**: each transfer is performed by a single arm, but either arm can be used. This means that two transfers can be performed simultaneously.
3) **Bilateral Handover**: each transfer is performed by both arms and consists of a pick followed by a handover followed by a place. Subsequent transfers can be pipelined as a pick can be performed for the next block while a place is performed for the current block. This is the standard surgeon training task in the FLS curriculum, though humans do not typically use pipelining.

We consider the Bilateral Handover variant of the peg-transfer task for the first time in an automated setting. The Bilateral Handover and Parallel Bilateral variants require precise coordination between two arms to avoid collisions and to efficiently execute handovers if needed.

We evaluate the system with three peg-transfer tasks. We use a bilateral dVRK [41, 42] with two large needle driver tools. We use a Zivid One Plus S RGBD camera mounted 0.5 m from the workspace. The camera has roughly a 50° vertical incline.

#### B. Notation

Let $\boldsymbol{q_p}$ be the robot's *physical* configuration, and define $C_p \subset \mathbb{R}^6$ to be the set of all possible configurations. The robot's *commanded* configuration is $\boldsymbol{q_p} \in C_c \subset \mathbb{R}^6$, which is equal to the encoder readings if the robot is allowed to come to a rest. This can differ from $\boldsymbol{q_p}$, because the encoders are located at the motors and away from the joints. The discrepancy between the joint configuration measured by the encoders and the true joint configurations is due to cabling effects.

Subscripts index specific joints in vectors, e.g., $\boldsymbol{q_p}^T = [q_{p,1} \cdots q_{p,6}]$. For ease of readability, we suppress the $_p$ and $_c$ subscripts when the distinction is not needed. We visualize the six joints $q_1, \cdots, q_6$ in Fig. 8. Let

$$\tau_t = (\boldsymbol{q_c}^{(0)}, \cdots, \boldsymbol{q_c}^{(t-1)}) \in \mathcal{T} \quad (1)$$

encode the prior trajectory information of the robot up to time $t$. We would like to estimate the function

$$f : C_c \times \mathcal{T} \to C_p \quad (2)$$

which is a dynamics model that maps the current command at time $t$ and prior state information to the current physical configuration of the arm. At execution time, we use a controller derived from $f$ by approximately inverting it for a desired output waypoint.

It is difficult to derive an accurate model for the dynamics $f$ that incorporates hysteresis, so we learn a parametric approximation using a deep neural network $f_\theta \approx f$ from a finite sequence of samples $\mathcal{D} = ((\boldsymbol{q_c}^{(t)}, \boldsymbol{q_p}^{(t)}))_{t=0}^{N}$.

## IV. Materials and Methods

### A. Robot Calibration using 3D Printed Fiducial Markers and Depth Sensing

We use both RGB and depth sensing to track the sphere fiducials and consider historical motions in the estimation of joint angles, which enables compensation for hysteresis and backlash-like effects. Furthermore, we design practical controllers using these models and benchmark the result of applying the proposed calibration procedure on a challenging peg transfer task.

We calibrate the robot by learning $f_\theta$ from data, which we can then use to more accurately control the robot. We start by sending a sequence of commands $\boldsymbol{q_c}$ and tracking the physical trajectories of 3D printed fiducial markers attached to the robot. We then convert the marker's positions to $\boldsymbol{q_p}$ using kinematic equations. After collecting a dataset of commands and physical configurations, we train a recurrent neural network to minimize the empirical risk for the mean square error (MSE) loss function.

It is challenging to accommodate additional encoders or sensors due to the confined space inside the surgical end effector. In this work, to estimate $\boldsymbol{q_p}$, we build on our prior work [10] and use an RGBD camera to track fiducials of colored spheres attached to the end effector (shown in Fig. 3). The two spheres on the shaft allows us to decouple the first three joints $(q_1, q_2, q_3)$ from the last three $(q_4, q_5, q_6)$ and thus to accurately estimate them excluding the cabling coupling. We design and place the four spheres on the jaw where they cannot overlap in the camera view within the working range of joints. Given the point clouds provided by the camera, we cluster the six groups of point sets by masking the color and size of each sphere. We then calculate each sphere's 3D position by formulating a least-squares regression.

We calculate six joint angles of the robot based on the dVRK kinematics. Since the end position of the tool shaft is only dependent on the first three joints, $(q_1, q_2, q_3)$, we can get the inverse function in combinations of the end position, which simply extend the two measured positions of spheres. We obtain the last three joints, $(q_4, q_5, q_6)$, by equating the rotation matrix of the end effector with the rotation matrix measured from the four spheres. The measurement accuracy is 0.32 mm in sphere detection and less than 0.01 radians in joint angle estimation.

We collect a training dataset $\mathcal{D}$ consisting of random, smooth motions of the arm, visualized in Fig. 4. This long trajectory is first collected by teleoperating the robot in random, continuous motions, *densely* recording the waypoints, and then replaying them with the fiducials attached. This enables us to collect ground truth information for trajectories that are not specific to peg transfer. During the process, we collect the configuration $\boldsymbol{q_p}$ estimated from the fiducials and commanded joint angles $\boldsymbol{q_c}$ to compile a dataset $\mathcal{D}$:

$$\mathcal{D} = ((\boldsymbol{q_c}^{(t)}, \boldsymbol{q_p}^{(t)}))_{t=0}^{N} \quad (3)$$

The dataset consists of $N=1355$ data points, and takes 18 minutes to collect. It takes under 1 min to train a model from the collected data. In the online execution phase, it takes less than 10 ms to predict the error and to compensate the motion error. In our prior work [10], we collect roughly the same amount of data, but from trajectories that consist of long, jerky motions resembling the peg transfer task, and sparsely sample waypoints from them. We find that dynamics models trained on those trajectories are prone to distribution shift errors if they are tested on tasks that consist of different motions from the training distribution. We hypothesize that we can address the distribution shift by collecting data $\mathcal{D}$ in a more randomized process, but while sampling at a finer-grained resolution.

From the camera view, we define the manipulator on the right as Patient Side Manipulator1 (PSM1) and on the left as PSM2. We repeat the data collection and calibrate both arms separately, as they have different cabling characteristics.

We sub-sample a portion of the previously described dataset with $N=1000$. Fig. 5 presents the desired and measured trajectory of each joint angle in both cases. We notice that the three

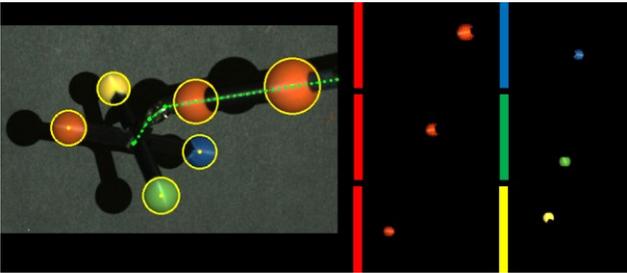

Fig. 3. dVRK calibration using 3D printed fiducial markers. We attach six fiducial spheres among which two of them on the tool shaft and four on the jaw. We convert the detected sphere positions into joint configurations $\boldsymbol{q_p}$ based on the dVRK kinematics. Left: The detected spheres are circled in yellow and the estimated joint configurations are overlaid with green dotted lines. Right: We cluster the point sets using color masking and the size of each sphere.

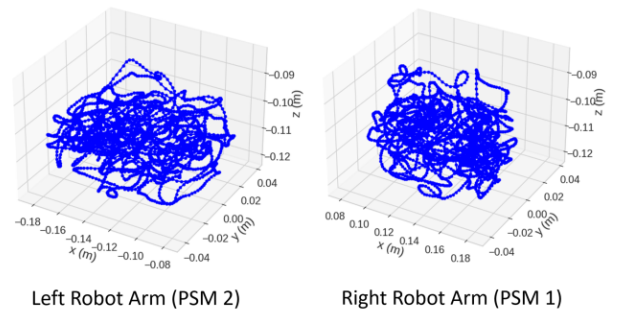

Fig. 4. Collecting random and smooth trajectories. We define the manipulator on the right as Patient Side Manipulator1 (PSM1) and on the left as PSM2. We repeat the data collection and calibrate both arms separately, as they have different cabling characteristics. We collect a dataset of random, smooth motions consisting of the configuration $\boldsymbol{q_p}$ estimated from the fiducials and commanded joint angles $\boldsymbol{q_c}$. The dataset consists of 1355 data points. These motions are not specifically optimized for the peg transfer task.

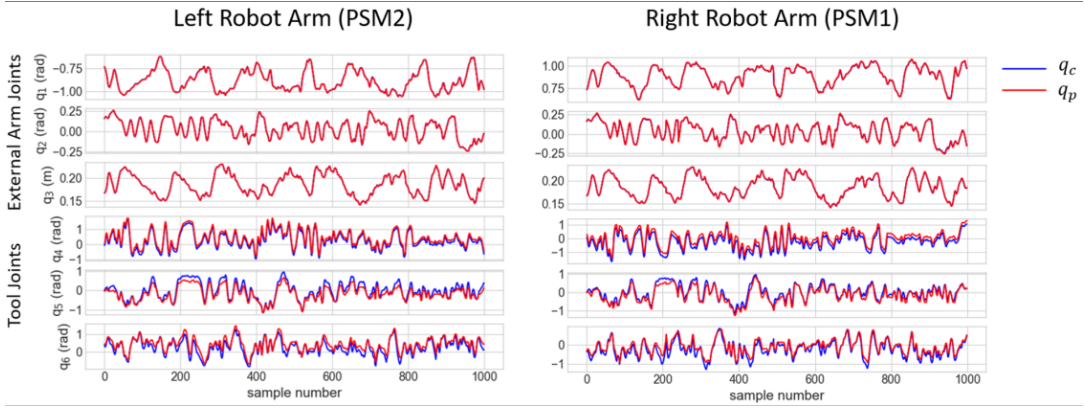

Fig. 5. Error identification of joint characteristics: We plot sub-sampled trajectories of the PSM1 and PSM2 from their training datasets which we use to calibrate the dVRK. Comparing the desired $q_c$ and measured $q_p$ configurations, we observe that the external arm joints (q1, q2, q3) are almost exactly overlaid and the wrist joints (q4, q5, q6) are the main source of the positioning error of the end effector, while the three joints from the external arm (q1, q2, q3) are relatively accurate. The quantitative results are shown in TABLE I.

TABLE I
STATISTICS OF ERROR IN THE TWO PATIENT SIDE MANIPULATORS (PSMS)

| | Left Robot Arm (PSM2) | | | | | | Right Robot Arm (PSM1) | | | | | |
|---|---|---|---|---|---|---|---|---|---|---|---|---|
| | $q_1$ (rad) | $q_2$ (rad) | $q_3$ (m) | $q_4$ (rad) | $q_5$ (rad) | $q_6$ (rad) | $q_1$ (rad) | $q_2$ (rad) | $q_3$ (m) | $q_4$ (rad) | $q_5$ (rad) | $q_6$ (rad) |
| **RMSE** | 0.0030 | 0.0036 | 0.00051 | 0.26 | 0.15 | 0.17 | 0.0020 | 0.0012 | 0.00015 | 0.16 | 0.17 | 0.21 |
| **SD** | 0.0017 | 0.0022 | 0.00030 | 0.024 | 0.095 | 0.099 | 0.0012 | 0.00068 | 0.000086 | 0.017 | 0.091 | 0.12 |
| **Max** | 0.0090 | 0.0110 | 0.0013 | 0.43 | 0.33 | 0.43 | 0.0054 | 0.0033 | 0.0045 | 0.25 | 0.32 | 0.45 |

In both cases, the first three joints have relatively small errors. Joint errors grow as joints get further from the proximal. PSM1 and PSM2 commonly have a small offset of $q_4$ and hysteresis with coupling of $q_5$ and $q_6$, which produces errors larger than 0.2 rad.

joints of the robot arm, $q_{p,1}$, $q_{p,2}$, and $q_{p,3}$, rarely contribute to the error compared to the last three joints from the statistics in TABLE I. We observe that the three joints of the surgical tool, $q_{p,4}$, $q_{p,5}$, and $q_{p,6}$, are repeatable and not affected by the arm joints. In addition, the last two joints are closely coupled, since $q_{p,5}$ synchronously moved with $q_{p,6}$ despite being commanded to be stationary, and vice versa. We hypothesize this occurs because these two joints have two additional cables that extend together along the shaft of the tool.

To estimate the configuration of the robot's joint angles $q_p$ without the fiducials attached, we train a function approximator $f_\theta: C_c \times \mathcal{T} \to C_p$, such that $f_\theta(q_c^{(t)}, \tau_t) = \hat{q}_p^{(t)} \approx q_p^{(t)}$. The model used is a Long Short-Term Memory (LSTM) [43] with 256 hidden units followed by two more hidden layers of 256 units. Five prior commands are supplied as input. We minimize the mean squared error (MSE) loss function between the model predictions $\hat{q}_p^{(t)}$ and ground-truth targets $q_p^{(t)}$ over the training dataset.

Once we train $f_\theta$, we would like to use it to accurately control the robot while compensating for the robot's cabling effects. At time $t$, the controller takes as input the target joint configuration $q_d^{(t)}$ and history-dependent input $\tau_t$ and computes joint configuration command $q_c^{(t)}$ to get the robot to that configuration. The controller (Alg. 1) iteratively refines the command based on the error relative to the target position. It evaluates the forward dynamics $f_\theta$ for a candidate command $q_c^{(t)}$ to obtain an estimate of the next configuration $f_\theta(q_c^{(t)}; \tau_t)$. Then, the algorithm modifies the input command to compensate for the error relative to the target position, executes the command, then updates the history $\tau_t$. This process repeats for $M$ iterations.

### B. Perception and Grasp Planning

In this section, we discuss how the calibrated and trajectory-optimized robot perceives the state of the pegs and blocks and generates a sequence of motions to perform peg transfer.

In our previous works [10, 12], we installed the depth camera to have a view perpendicular to the task board, which allows the point sets on the blocks to be obtained by cropping a depth range. To enable perception from a camera in a greater variety of poses, we use a point set registration algorithm, Iterative Closest Point (ICP) [44], which is simple and computationally efficient.

At the beginning of the task, we register the captured point clouds from the known 3D model of the pegboard to obtain the transformation matrix ${}^C_p T$ from the camera to the pegboard. Then, we obtain the 12 peg positions by cropping a height range

---

**Algorithm 1** Control Optimization Algorithm

**Require**: Target configuration $q_d^{(t)}$, state estimator $f_\theta$, number of iteration M, learning rate α

1: $q_c^{(t)} \leftarrow q_d^{(t)}$
2: **for** $j \in \{1, ..., M\}$ **do**
3:     $\Delta_j \leftarrow q_d^{(t)} - f_\theta(q_c^{(t)}; \tau_t)$     Estimate error
4:     $q_c^{(t)} \leftarrow q_c^{(t)} + \alpha \Delta_j$     Adjust based on error
5: **end for**
6: **return** $q_c^{(t)}$

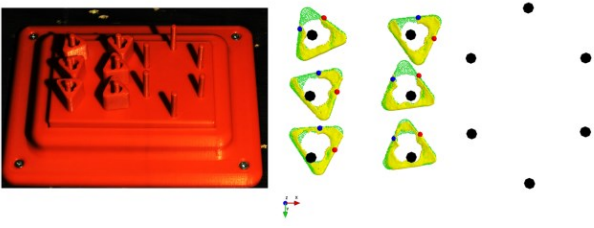

Fig. 7. Perception pipeline using depth sensing: We use point set registration to get poses of the pegboard and blocks. We perform a sequence of RGBD capturing, point set clustering, and registration of the block between each transfer motion. Left: Peg transfer environment from the inclined camera view. Right: Point sets of the detected pegs (black) and blocks (yellow), the registered 3D block model (green), the planned grasping point of the left arm (blue) and right arm (red). We capture point sets from the inclined camera and change the view perspective to top-down for better visualization.

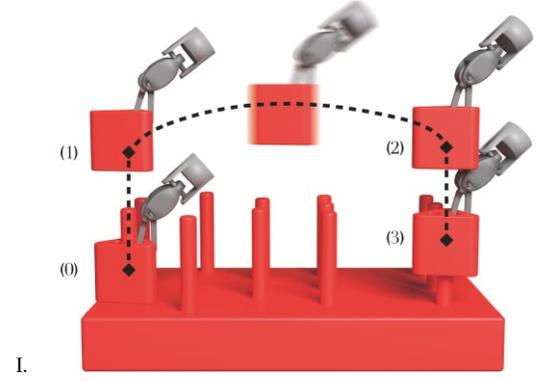

Fig. 6. Optimized trajectory segments for a transfer motion. Each peg transfer consists of 4 waypoints and the three time-optimized splines between. The motion starts at waypoint (0) and ends at waypoint (3) with zero velocity. At waypoints (1) and (2), the splines are connected to ensure $C^1$ continuity, where the velocities are selected to minimize execution time.

with respect to the base frame of the pegboard. Given the desired peg position and the known height of the block, we cluster a point set on the block to be transferred and get a transformation from the camera to the block ${}^{c}_{b}T$ using ICP. Since the transformation from camera to robot ${}^{c}_{r}T$ is known from the robot calibration procedure, we finally obtain the block pose with respect to the robot's base frame. During the task, we perform a sequence of RGBD capturing, point set clustering, and registration of the block between each transfer motion. These process takes an average of 241 ms per each block.

After detecting the blocks, we find a best grasp among a total of the six potential poses per block. We subdivide each block edge into two potential grasps and pre-calculate the six grasps for the 3D block model. See Fig. 7 for details. Given the block poses, we transform these grasping poses using ${}^{c}_{b}T$ and calculate the distance from the peg to find the farthest point among the poses that allows jo int limits. We avoid the grasping point behind a peg to decrease the chances of collision.

### C. Non-Linear Trajectory Optimization

In this section, we describe a trajectory time-optimization to improve the speed of task performance. In prior formulations [45, 46] of pick-and-place trajectory optimization, a time-minimized trajectory is found by discretizing the trajectory into a sequence of waypoints and formulating a sequential quadratic program that minimizes the sum-of-squared-acceleration or sum-of-squared distance between the waypoints. We observe that this prior formulation, while versatile enough for the peg transfer tasks, can be simplified to minimizing a sequence of splines defined by up to four waypoints, and relying on the kinematic design of the robot to avoid collisions. By reducing to four waypoints, we trade off some optimality of motion for the reduced computational complexity of the trajectory optimization.

The objective of this optimization is to minimize the trajectory motion time. The peg transfer task consists of 12 block transfers, each of which sequences motions that lift up, translate horizontally, and lower down. Due to the small clearance between blocks and pegs, the lifting motion for a block should be parallel to the peg until the block clears the top of the peg. Lowering the block onto a peg requires a similar parallel motion (Fig. 6).

When computing a trajectory for the robot, we convert end-effector poses (e.g., for the pick location) to robot configuration using an inverse kinematic (IK) calculation. General IK solvers typically treat this problem as an iterative optimization problem and thus may take many iterations to converge to a solution. Due to the kinematic design of the dVRK robot, we observe that a closed-form analytic solution exists that allows for fast IK solutions.

We apply Pieper's method [47] to find a closed-form IK solution. Pieper proved that an analytic solution always exists for a 6 DoF manipulator under the condition that the last three axes mutually intersect. Many laparoscopic surgical robots have a special mechanical constraint, called Remote Center of Motion (RCM), to constrain movement at a point where the patient's incision is positioned. We apply this method by inversely reassigning the coordinates as shown in Fig. 8. We consider the end effector as a base frame. This method can be applied to any 6 DoF surgical manipulator with RCM motion constraint.

By using the closed-form IK solution, we further speed up trajectory computation time. Our IK solution achieves an average compute time of $0.8\pm0.2$ ms, while the numerical optimization method, which uses 1,000 random joint configurations, takes $116.3\pm12$ ms. We use the SciPy module in Python for the comparison.

We define $f_{\text{IK}}: SE(3) \to \mathbb{R}^6$ as the IK function, w here the input is the pose of the end-effector and the output is the robot configuration:

$$f_{\text{IK}}\left(\begin{bmatrix} \boldsymbol{R} & \boldsymbol{t} \\ \boldsymbol{0} & \boldsymbol{1} \end{bmatrix}\right) = [q_1 \quad q_2 \quad q_3 \quad q_4 \quad q_5 \quad q_6]^T$$

and from Pieper's method, we derive the following:

$$\begin{aligned}
\boldsymbol{t}^{inv} &= -\boldsymbol{R}^T \boldsymbol{t} \\
q_6 &= arctan2(t_x^{inv}, t_z^{inv} - L_4) \\
\text{p} &= -L_3 + \sqrt{(t_x^{inv})^2 + (t_z^{inv} - L_4)^2} \\
q_3 &= L_1 - L_2 + \sqrt{(t_y^{inv})^2 + p^2} \\
q_5 &= arctan2(-t_y^{inv}, p)
\end{aligned}$$

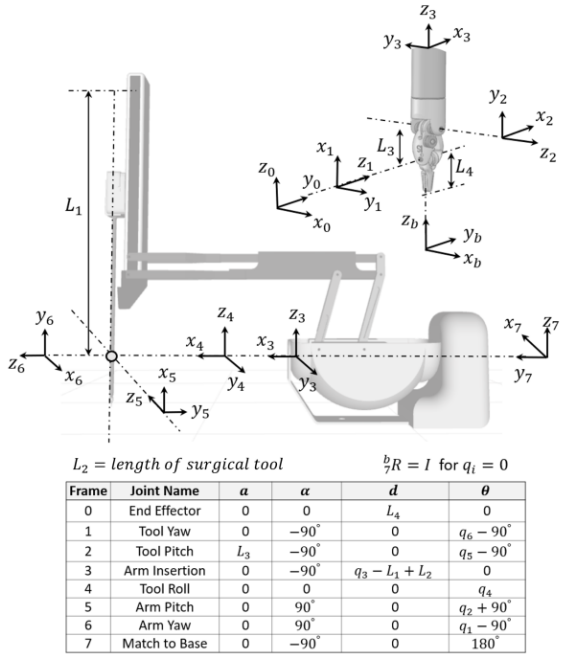

Fig. 8. Coordinate frames for closed-form inverse kinematics: We follow the modified Denavit-Hatenberg convention, by inversely re-assigning coordinates to calculate a closed-form IK solution according to Pieper's method. We consider the end effector as a base frame.

$$R_3^b = \begin{bmatrix} sinq_5 \cdot sinq_6 & -cosq_6 & cosq_5 sinq_6 \\ cosq_5 & 0 & -sinq_5 \\ cosq_6 \cdot sinq_5 & sinq_6 & cosq_5 \cdot cosq_6 \end{bmatrix}$$

$$R_7^3 = R_3^{b^T} \cdot R^T$$
$$q_2 = sin^{-1}(R_7^3[3,2])$$
$$q_1 = arctan2(-R_7^3[3,1], R_7^3[3,3])$$
$$q_4 = arctan2(R_7^3[2,2], R_7^3[1,2])$$

Here we use the notation $\mathbf{R}[1,2]$ to refer to the coefficient at row 1 and column 2 of the matrix $\mathbf{R}$, and subscripts in $t_x$ and $t_y$ to refer to the respective coefficients of the vector $\mathbf{t}$. The scalars $L_1$, $L_2$, $L_3$, and $L_4$ represent physical measurements of the dVRK, visualized in Fig. 8.

We propose an optimization based on a cubic spline interpolation in the joint space to compute a fast, smooth, and feasible trajectory to do the peg-transfer motions. To compute the spline, we define a cubic polynomial based on the configuration $\mathbf{q}$ and configuration-space velocity $\mathbf{v}$ at the start $(q^{(i)}, v^{(i)})$ and end $(q^{(i+1)}, v^{(i+1)})$ of each motion segment $i$. We then combine a sequence of splines to traverse through the whole peg-transfer motion. At each point where one spline ends and the next starts, we ensure $C^1$ continuity by setting the configuration and velocity of the end point. We define a spline $f_{spline}$ as a function of time $t$, segment parameterized start and end points, and duration of the spline segment $t_i$ as follows:

$$f_{spline}\left(t; \begin{bmatrix} q^{(i)} \\ v^{(i)} \end{bmatrix}, \begin{bmatrix} q^{(i+1)} \\ v^{(i+1)} \end{bmatrix}, t_i\right) = at^3 + bt^2 + ct + d$$

where,

$$\mathbf{a} = \frac{v^{(i+1)} + v^{(i)}}{(t_i)^2} + \frac{2(q^{(i)} - q^{(i+1)})}{(t_i)^3}$$
$$\mathbf{b} = \frac{1}{2}\frac{v^{(i+1)} - v^{(i)}}{t_i} - \frac{3}{2}\mathbf{a}t_i$$
$$\mathbf{c} = v^{(i)}$$
$$\mathbf{d} = q^{(i)}$$

For a peg-transfer motion we define a trajectory by 4 waypoints (and therefore 3 segments): (0) the pose in which the robot is initially grasping the block on its starting peg, (1) the pose with the block lifted to clear its starting peg, (2) the pose with the block vertically over its ending peg, and (3) the pose with the block lowered onto the ending peg and about to release its grasp (Fig. 6).

The configurations of these waypoints are all defined by the IK solution and denoted by $(q^{(0)}, q^{(1)}, q^{(2)}, q^{(3)})$. We denote the velocities at these waypoints by $(v^{(0)}, v^{(1)}, v^{(2)}, v^{(3)})$. The velocity at waypoints (0) and (3) is zero. The velocities at (1) and (2) are in the direction of lifting and lowering respectively, with their magnitude optimized through gradient descent. We let $t_i$ denote the duration of segment $i$. For a single transfer motion, we thus have the following trajectory sequence:

$$f_{transfer}\left(t; q^{(0)}, q^{(1)}, q^{(2)}, q^{(3)}, v^{(1)}, v^{(2)}, t_0, t_1, t_2\right) =$$
$$\begin{cases} f_{spline}\left(t; \begin{bmatrix} q^{(0)} \\ 0 \end{bmatrix}, \begin{bmatrix} q^{(1)} \\ v^{(1)} \end{bmatrix}, t_0\right) & if\ t < t_0 \\ f_{spline}\left(t - t_0; \begin{bmatrix} q^{(1)} \\ v^{(1)} \end{bmatrix}, \begin{bmatrix} q^{(2)} \\ v^{(2)} \end{bmatrix}, t_1\right) & if\ t_0 \le t < t_0 + t_1 \\ f_{spline}\left(t - t_0 - t_1; \begin{bmatrix} q^{(2)} \\ v^{(2)} \end{bmatrix}, \begin{bmatrix} q^{(3)} \\ 0 \end{bmatrix}, t_2\right) & if\ t_0 + t_1 \le t \end{cases}$$

We compute the direction of the velocity $\hat{v}$ at (1) and (2) in joint space by computing the direction of the motion in end-effector space and transforming it through the inverse of the Jacobian of the robot at the corresponding configuration, thus:

$$\hat{v}^{(1)} = \left(J^{(1)}\right)^{-1}\hat{z}$$

where $\hat{z}$ is the direction to lift for (1), and the direction to lower for (2). In practice, we orient the peg board such that the pegs are aligned to the $z$-axis.

To compute the duration of the splines $t_0$, $t_1$, $t_2$, we compute spline coefficients for integer multiples of 0.01 s, and select the shortest duration that satisfies all joint velocity and acceleration limits. The maximum velocity of the spline occurs at the root of the second derivative (when the spline's coefficients $6\mathbf{a} + 2\mathbf{b} = 0$). The maximum acceleration occurs at either end of the spline (thus $2\mathbf{b}\ or\ 6\mathbf{a}t + 2\mathbf{b}$).

To compute the magnitude of the velocity $\lambda$ at (1) and (2), and thus get the velocity $v^{(1)} = \lambda^{(1)}\hat{v}^{(1)}$, we iteratively take gradient steps according to a cost function $f_{cost}$ defined to be $t_0 + t_1 + t_2$ that minimizes the time for each segment of the trajectory. This algorithm computes the time-optimized trajectories and is shown in Alg. 2.

## D. Time-Optimal and Jerk-Minimizing Trajectory Refinement

The spline-based optimization computes a trajectory quickly, but it is by construction, not time-optimal. A time-optimal trajectory at any moment in time will be at a joint limit (e.g., velocity and acceleration). However, in the spline-based optimization, we only compute a maximum of 3 splines, meaning that velocities and accelerations will tend to smoothly vary instead of plateauing at a maximum.

To then speed up the trajectory further, we use the spline-based optimization to warm-start [46] a time-optimizing sequential quadratic program (SQP). This SQP-based optimization adds additional spline segments while enforcing motion limits, and successively shortens the time horizon until it is detected as infeasible. As such, this SQP formulation follows closely to that of a grasp-optimized motion planning [45], though we increase the tolerances to speed up the computation, and leave grasp optimization to future work.

We compute an SQP to refine each segment from the spline-based optimization. Each SQP optimizes a non-convex QP of the following form:

$$\operatorname*{argmin}_{q^{[0..H]}, v^{[0..H]}, a^{[0..H]}} \sum_{t=1}^{H-1} |(a^{(t)} - a^{(t-1)})/T|^2$$

subject to
$q_{t+1} = q_t + v_t T + a_t T^2/2 \quad \forall t \in [0...H]$
$v_{t+1} = v_t + a_t T \quad \forall t \in [0...H]$
$q_t, v_t, a_t$ in joint limits $\quad \forall t \in [1...H]$
$q_t$ avoids obstacles $\quad \forall t \in [1...H]$

where $T$ is a fixed time interval between each waypoint, and $H$ is the number of waypoints in the trajectory. We start with $T \times (H + 1)$ equal to the duration of the spline computed in the spline-based optimization. The objective minimizes the sum-of-squared jerks computed as the change in acceleration over time. The first pair of constraint sets enforces consistency of the dynamics between configuration $q$, velocity $v$, and acceleration $a$ and results constant-acceleration spline between each waypoint, matching the splines from before. The third constraint set ensures that the robot remains within the configuration, velocity, and acceleration limits. The last constraint set avoids obstacles, e.g., block and peg contact.

Since the obstacle avoidance constraints are non-convex, we linearize them at each the iteration of the SQP solver, using the previous solution iteration as the linearization point. This linearization takes the following form:

$$Jq = p + Jq_{prev} - f_{FK}(q_{prev})$$

where $p$ is the center coordinate $(x,y)$ of the vertical channel through which the peg must be lifted, $f_{FK}$ is the forward-kinematics function, and $J$ is a Jacobian relating the change in joint angle to the change in end-effector pose. We assume that the motion is vertical through $z$; thus, the constraint only enforces $x$ and $y$ position. We box-bound this constraint with a small tolerance and add a linear penalty using slack variables.

Depending on which motion we are optimizing, we vary this SQP by adding constraints on the boundary conditions. For example, to lift a block off of a peg, we constrain $q_0$ to the grasp

---

**Algorithm 2** Optimize Peg Transfer Splines

**Require**: Waypoints $Q = (q^{(0)}, q^{(1)}, q^{(2)}, q^{(3)})$, directions $V = (\hat{v}^{(1)}, \hat{v}^{(2)})$, step rate $\beta$, maximum iterations max_iter

1: $[\lambda^{(0)} \ \lambda^{(1)}]^T \leftarrow 0$          Initial guess
2: $c_{prev} \leftarrow f_{cost}(Q, V, [\lambda^{(1)} \ \lambda^{(2)}]^T)$
3: **for** iter $\in \{1, ..., \text{max\_iter}\}$ **do**
4:     $\begin{bmatrix}\lambda^{(1)}\\ \lambda^{(2)}\end{bmatrix} \leftarrow \begin{bmatrix}\lambda^{(1)}\\ \lambda^{(2)}\end{bmatrix} - \beta \nabla f_{cost}\left(Q, V, \begin{bmatrix}\lambda^{(1)}\\ \lambda^{(2)}\end{bmatrix}\right)$
5:     $c_{curr} \leftarrow f_{cost}(Q, V, [\lambda^{(1)} \ \lambda^{(2)}]^T)$
6:     **if** $|c_{curr} - c_{prev}| <$ tolerance **then**
7:        break
8:     **end if**
9:     $c_{prev} \leftarrow c_{curr}$
10: **end for**
11: **return** $\lambda^{(1)}, \lambda^{(2)}$

---

configuration, $v_0 = 0$, and $q_H$ to the configuration over the peg. We only constrain $v_H$ to be within the joint limits, thus allowing the optimization to have high-velocity at the end of the motion, where it would then connect to the next spline. For the next spline, we then set its $v_0$ to be the optimized $v_H$ from the previous trajectory. We apply a similar approach to dropping motions using appropriately set boundary conditions (i.e., unconstrained starting velocity, stopping at the end of the motion).

To convert this into a time-optimization, after solving the SQP we shorten $H$, and repeat the process, until the solver detects the SQP as infeasible. We use the shortest $H$ for which a solution was found. As this process starts with a trajectory duration of $T \times H$ from the spline, if the first SQP is infeasible, we use trajectory from spline-based optimization.

In practice, setting $T$ to the controller frequency of the dVRK (10 ms), will lead to the shortest trajectory, as it will changes in accelerations to be applied at the most rapid interval. However, as this can lead to a large $H$, it can slow down the computation. We thus set a value of $T$ and $H$ that balances the compute time and trajectory time, while keeping $T$ as an integer multiple of the controller frequency. We then interpolate the solution to the controller frequency using the dynamics equations from the SQP. The trajectory planning is done online and the computation takes 103 ms per trajectory in average.

## E. Trajectory Analysis

To analyze the joint and end-effector velocity, acceleration, and jerk, we record the joint encoders throughout each trial, and follow and extend a method outlined by Todorov and Jordan [48]. This method formulates a quadratic polynomial for each segment of the measured trajectory and includes velocities and accelerations for each joint measurement. It then computes a minimization of the velocity and accelerations across all segments using the nonlinear simplex method. We modify this method to minimize a Huber loss on the dynamics equations that explain the measurements, and solve using a quadratic program solver. This QP takes the form:

$$\operatorname*{argmin}_{v^{[0..H_m]}, a^{[0..H_m]}} \sum_{t=0}^{H_m-1} 2(u_t^q)^2 - u_t^q + r_t^q + s_t^q + 2(u_t^v)^2 - u_t^v + r_t^v + s_t^v$$

$$\text{subject to} \quad q_{t+1} = q_t + v_t T_m + \frac{a_t T_m^2}{2} + u_t^q + r_t^q - s_t^q \quad \forall t \in [0 \dots H_m]$$
$$v_{t+1} = v_t + a_t T_m + u_t^q + r_t^q - s_t^q \quad \forall t \in [0 \dots H_m]$$
$$r_t^q \geq 0, \; s_t^q \geq 0 \quad \forall t \in [0 \dots H_m]$$
$$r_t^v \geq 0, \; s_t^v \geq 0 \quad \forall t \in [0 \dots H_m]$$

where $H_m$ is the number of configurations measured, $T_m$ is the time interval between measurements, and $q_t$ is the measured configuration at time step $t$. The slack variables $u_t^q$, $u_t^v$ form the quadratic part of the Huber loss, while $r_t^q$, $r_t^v$, $s_t^q$, and $s_t^v$ are the linear part of the Huber loss. After solving, we then analyze the velocity and acceleration variables independently, and compute jerk as the difference between accelerations divided by $T_m$. We analyze automated trials using this method too, as the measurements of executed trajectories may not precisely match the computed values.

## V. Physical Experiments

### A. Robot Experiments

For each experiment, we measure robot autonomous performance under 3 conditions:

1) *Uncalibrated, Unoptimized*: This uses the default robot PID controller that uses encoder estimates to track trajectories and performs no trajectory optimization.
2) *Calibrated, Unoptimized*: This uses the calibration procedure from [10] but does not optimize trajectories for pick and place and handover. We hypothesize the calibration procedure will increase the accuracy of the system, but may not necessarily change its speed.
3) *Calibrated, Optimized*: This uses both the robot calibration procedure and the trajectory optimization techniques. We do not consider trajectory optimization without calibration, as high-speed collisions may damage instruments and we do not expect optimization to affect the accuracy of the uncalibrated model. We hypothesize that this method will be as accurate as Calibrated, Unoptimized, but result in much faster motions.

### B. Human Subjects Protocol

After obtaining IRB protocol approval, we enrolled 10 human subjects: 6 males, including one surgical resident, and 4 females with an average age of 26.2.

Except for the surgical resident, they are undergraduate/graduate students or postdoctoral researchers at University of California Berkeley. None of the human subjects in the volunteer group have experience using the da Vinci nor performing robot teleoperation. The surgeon is co-author Dr. Danyal Fer. He has 4 years of general surgery experience, and has performed over 900 open and minimally invasive procedures. The human subjects were asked to teleoperate 2 standard master handles to control the two surgical arms and view the workspace through an endoscopic stereo camera.

We asked 9 volunteers to complete six 12 block transfer trials for each of the 3 variants of the peg transfer tasks described in Fig. 2 for a total of 1944 trials. Prior to the experiment, we give volunteers a short (5 minutes) instruction on how to control the da Vinci surgical robot. The recording is started at the beginning of the first trial without practice. We record the angle and velocity of each joint. The sampling period of the recording is 10 ms. The experiment takes about 80 minutes per volunteer.

The experimental protocol was approved by the Institutional Review Board (IRB) and the Committee for Protection of Human Subjects (CPHS) of the University of California, Berkeley (2021-02-14049).

### C. Experiment Variables and Hypothesis

For each of the three peg transfer versions, we control the independent variable of robot operator, varying between human subject (*Volunteer, Surgeon*) and autonomous control method (*Uncalibrated Unoptimized, Calibrated Unoptimized, Calibrated Optimized*). For each trial, we record dependent variables of transfer success rate, transfer time, distance traveled, and collisions for each robot operator. We test the following hypotheses:

H1) The Calibrated Optimized and Calibrated Unoptimized versions will achieve similar transfer success rates that are on par with the surgeon and greater than the volunteers.

H2) The Calibrated Optimized robot will have a lower mean transfer time than the surgeon who will in turn have a lower mean transfer time than the volunteers. This may particularly apply for the bilateral peg transfer tasks, because the robot can easily parallelize its transfers.

H3) The Calibrated Optimized robot trajectories will have fewer collisions per trial than the Uncalibrated robot trajectories and the human trajectories.

H4) The Calibrated Optimized robot will have a lower distance traveled per trial than the surgeon. The surgeon will cause the robot to have a lower distance traveled than the volunteers.

### D. Results

We present and discuss results for the unilateral, parallel bilateral, and bilateral handover cases in TABLE II, TABLE III, and TABLE IV, respectively. Each case involves running the three robot conditions: (1) *Uncalibrated, Unoptimized*, (2) *Calibrated, Unoptimized*, and (3) *Calibrated, Optimized*, plus the surgical resident (*Surgeon*) and inexperienced teleoperators (*Volunteers*) for 10 trials each. When presenting the Volunteers results, we average all results from the 9 volunteers (648 transfers per each variant). We present aggregated trajectory statistics such as distance traveled and average acceleration in Fig. 9 and Fig. 10.

1) *Unilateral (One Arm) Peg Transfer*

Using only one arm, the surgeon has the fastest mean transfer time of 4.7 s, with a perfect 120/120 success rate. The calibrated robot also attains perfect 120/120 success rates, matching the surgeon's performance. Volunteers have a relatively high success rate of 95.9 %, but require an average of 12.3 s per transfer attempt. The *Volunteers* and the *Surgeon* are able to recover some of the failures to increase their success rate. We count the recovered failures as success. The increased success rate by recovery is 16/636 = 2.5 % in Volunteers.

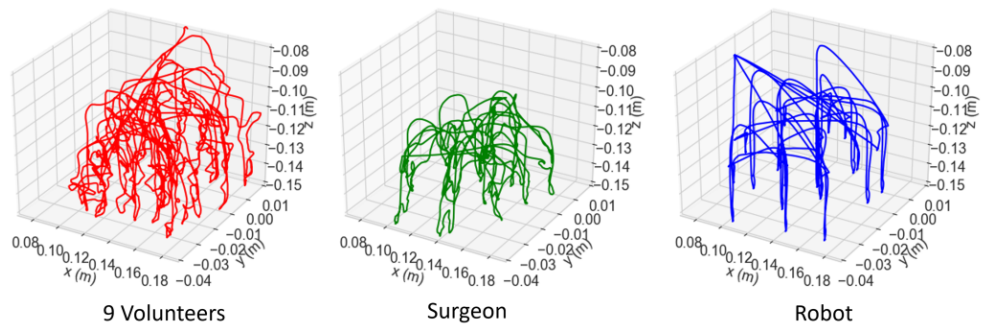

Fig. 9. Position trajectory examples. These examples show position trajectories of surgical tool-tip during one completion of trial (12 block transfers). *Volunteers* tend to take long and winding paths. This is because they have several trials and errors until a successful pick-up or placing. The *Surgeon* makes a relatively smoother, but still wobbly trajectory. The *Surgeon* often has an overshoot that passes through the target peg and returns to it. The *Robot* creates a smooth trajectory which is consistent during all trials.

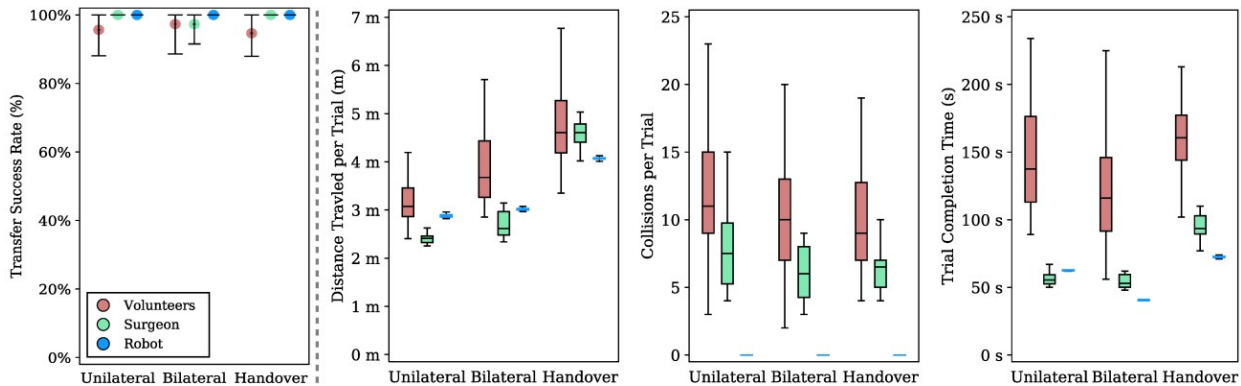

Fig. 10. Performance comparison. We analyze the performance of 9 volunteers, a surgeon with more than 150 hours of professional experience with the surgical robot, and the automated robot, in terms of transfer success rates, tool-tip distance traveled per trial, number of collisions per trials, and completion time per trial. For each task variant, the statistics are calculated from 54 trials by the *Volunteers*, 10 trials by the *Surgeon*, and 10 trials by the robot (*Calibrated, Optimized*). The leftmost plot shows the mean success rate, with one standard deviation error bar, while the plots to the right utilize a traditional boxplot. On average, all three succeed on the vast majority of transfers, but the robot is more reliable than all humans for the bilateral variant. While distance traveled is comparable for the surgeon and the robot, in terms of collisions the robot is superior for all three task variants, and for completion time in the bilateral and handover variants. Additionally, the robot exhibits the lowest variance for all performance measures. (Total completion time: $145.8\pm44.7$ vs $56.7\pm6.0$ vs $62.6\pm0.7$ s for unilateral, $129.7\pm56.5$ vs $56.7\pm11.2$ vs $40.4\pm0.5$ s for bilateral, and $181.8\pm61.9$ vs $94.8\pm10.0$ vs $72.5\pm1.3$ s for handover) This suggests that it also achieves superhuman consistency in its motions.

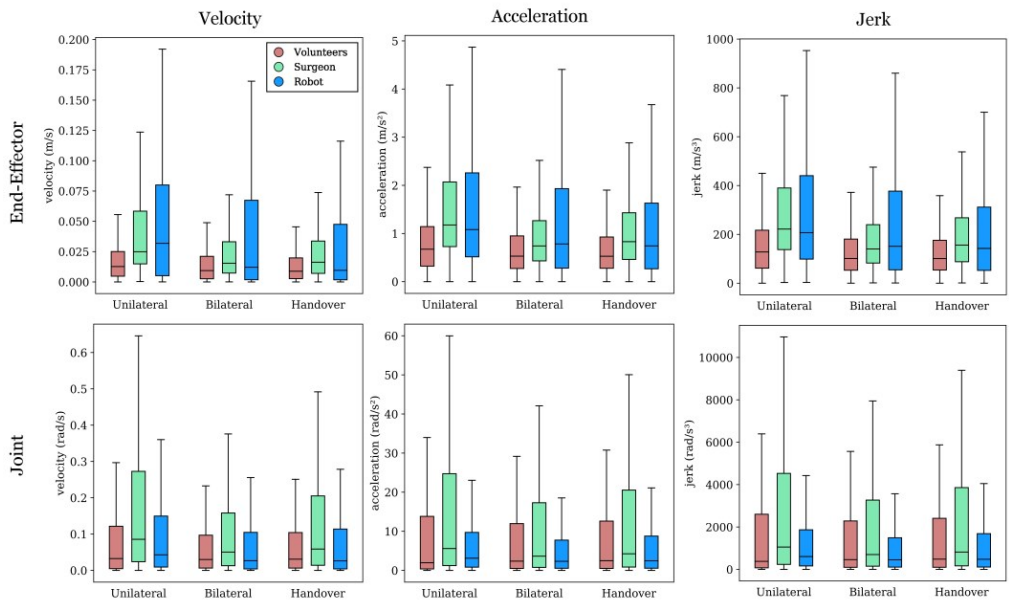

Fig. 11. Comparison of velocity, acceleration, and jerk of volunteers, surgeon, and the robot. The 9 volunteers, as they are new to the system, tended to operate at below the capabilities of the robot, while the surgeon and the robot pushed the limits. The surgeon operated at higher velocities, with higher accelerations and jerk than the volunteers. While the surgeon places the most demands on the joints, the robot has the fastest motion in the task-relevant end-effector space.

TABLE II
UNILATERAL PEG-TRANSFER TASK EXPERIMENTS

| Method | Mean Transfer Time (s) | Success / Attempts | Success Rate (%) |
|---|---|---|---|
| 9 Volunteers | 12.3 | 610/636 | 95.9 |
| Surgeon | **4.7** | **120/120** | **100.0** |
| Robot Uncalibrated, Unoptimized | 5.6 | 26/75 | 34.7 |
| Robot Calibrated, Unoptimized | 5.7 | **120/120** | **100.0** |
| Robot Calibrated, Optimized | 5.2 | **120/120** | **100.0** |

The calibrated robot and the surgeon both achieve a 100 % success rate on 10 trials. We define the mean transfer time as the completion time divided by the number of block transfers. Despite trajectory optimization, the surgeon is still 10.6 % faster than the robot. Factors that delay the automated robot include computing registration of the block and grasping pose (0.098 s) and calculation of online trajectory planning (0.103 s). These factors are not optimized in this paper, which suggesting that there is still further room for improvement. The actual motion execution speed of the robot without perception is slightly faster than the surgeon.

TABLE III
PARALLEL BILATERAL PEG-TRANSFER TASK EXPERIMENTS

| Method | Mean Transfer Time (s) | Success / Attempts | Success Rate (%) |
|---|---|---|---|
| 9 Volunteers | 10.9 | 626/642 | 97.5 |
| Surgeon | 4.8 | 115/118 | 97.5 |
| Robot Uncalibrated, Unoptimized | 3.6 | 37/68 | 54.4 |
| Robot Calibrated, Unoptimized | 3.5 | **120/120** | **100.0** |
| Robot Calibrated, Optimized | 3.0 | **120/120** | **100.0** |

The Calibrated, Optimized robot is slightly more successful and 60 % faster than the surgeon. Unlike the unilateral case, even the unoptimized trajectories are faster than the surgeon, whose mean transfer time (4.8 s) is similar to the time in the unilateral case (4.7 s). This suggests that parallelism is easier for the robot than for the surgeon.

TABLE IV
BILATERAL HANDOVER PEG-TRANSFER TASK EXPERIMENTS

| Method | Mean Transfer Time (s) | Success / Attempts | Success Rate (%) |
|---|---|---|---|
| 9 Volunteers | 15.5 | 599/632 | 94.8 |
| Surgeon | 7.9 | **120/120** | **100.0** |
| Robot Uncalibrated, Unoptimized | 6.3 | 14/79 | 17.7 |
| Robot Calibrated, Unoptimized | 6.1 | **120/120** | **100.0** |
| Robot Calibrated, Optimized | **6.0** | **120/120** | **100.0** |

We observe that the Calibrated, Optimized method is able to outperform the surgeon in terms of success rate and speed. The mean transfer time for the surgeon increased (i.e., slowed down) 68 % from the unilateral task, while the mean transfer time for the Calibrated, Optimized method only increased by 15 %. The results here, coupled with those from the parallel bilateral experiments, suggest that the robot is better able to parallelize subtasks than the human.

The fastest reported times for an autonomous surgical robot on this task and setup with a comparable success rate is 13.6 s per transfer [10], suggesting a substantial improvement of 262 %. Within this study, the trajectory optimization procedure is able to reduce mean transfer time by 10 %. However, the Calibrated, Optimized method is still slower than the *Surgeon*. Factors that delay the automated robot include computing registration of the block and grasping pose (0.098 s) and calculation of online trajectory planning (0.103 s). After moving 6 blocks from the left-half side to the right-half side, the robot moves out of the workspace for another RGBD capturing. This takes more than 3 s per trial. The actual motion execution speed of the robot without perception is slightly faster than the surgeon. These factors are not optimized in this paper, which suggests that there is still further room for improvement.

2) *Parallel Bilateral (Two Arms) Peg Transfer*

In this case, we find that the Calibrated, Optimized method is 60 % faster than the surgeon in mean transfer time, and is better than the surgeon in success rate (120/120 versus the surgeon's 115/118). This task was also studied in prior work [10], which reports a success rate of 78.0 % and a mean transfer time of 5.7 s. In contrast, this work improves the success rate by over 28 % and speeds up the mean transfer time by 19.0 %. Volunteers increase the success rate using recovery by 22/642 = 3.4 % and the surgeon by 4/118 = 3.4 %. The *Surgeon* experienced 7 failures during bilateral transfers and recovered in 3 cases. In the remaining 4 cases, the *Surgeon* is not able to recover the block because the block falls out of the workspace of the robot.

3) *Bilateral Handover Peg Transfer*

In the bilateral handover case, we find that the Calibrated, Optimized method has the highest success rate and the fastest mean transfer time (6.0 s). The Calibrated, Optimized method outperforms the surgeon by 31.7 % in mean transfer time. We attribute the faster speed in this setting to the surgeon's observation that ensuring the handover step works correctly is difficult and requires time and care. In addition, the way the robot does the automated handover is slightly different and does not involve a significant rotation of the wrist joints. While the surgeon rotates the block with the first gripper into a horizontal orientation to grasp it on the opposite side with the second gripper, the robot maintains the vertical orientation of the block and grasps the same side of the block with both grippers during handover. To our knowledge, this is the first automation study of the bilateral handover peg-transfer task.

4) *Trajectory Analysis*

Fig. 10 presents a comparison of the *Volunteers*, *Surgeon*, and the robot under the *Calibrated Optimized* condition. We do not show the uncalibrated robot because it preforms significantly worse. We observe that the robot and the *Surgeon* both have near perfect success rates on all tasks. The *Surgeon* is faster than the automated procedure on the unilateral task, but is slower on average on the parallel bilateral and bilateral handover tasks. The robot is faster than the fastest volunteer in all trials. We define collision as the two robot arms colliding during hand off or the robot arm/the block hitting the peg on the sides during the task. In most cases, collisions are recovered and do not result in failures. The robot never has collisions, whereas both the surgeon and volunteers have a median of 6 and 8 collisions per trial respectively on all tasks. We also measure the distance traveled by the end effectors during the tasks. The volunteers tip motion is significantly longer than that of the surgeon and the robot on all tasks. The surgeon mo tion is slightly shorter than the robot on the unilateral task, roughly the same in the bilateral task and over 12.5 % larger in median on the handover task.

Fig. 11 presents the velocity, acceleration, and jerk during task execution for both the joints and end eff ector, using the method from Todorov and Jordan [48]. We observe that the volunteers operate at a lower velocity, acceleration, and jerk than the surgeon and the robot. This suggests that familiarity with system limits and capabilities helps the surgeon execute faster motions. Interestingly, while the surgeon shows higher velocity, acceleration, and jerk on the joints, the robot has higher velocity, acceleration, and jerk in the end-effector. This suggest that the robot is able to achieve higher performance in the task-relevant end-effector, potentially reducing wear on the joints.

## VI. Conclusion

This paper significantly extends our prior work on automating surgical subtasks, in particular the Fundamentals of Laparoscopic Surgery peg-transfer task and present results for three variants: unilateral, bilateral without handovers, and bilateral with handovers.

For the most difficult variant of peg transfer (bilateral with handovers) over 120 trials, the surgeon achieves success rate 100.0 % with mean transfer time of 7.9 s. The robot achieves success rate 100.0 % with mean transfer time of 6.0 s. On all three variants of the task, the Calibrated Optimized and Calibrated Unoptimized robot achieved comparable success rates to the surgeon and outperformed the volunteers, confirming Hypothesis H1. In the bilateral variants of the task, the robot was faster than the surgeon, confirming Hypothesis H2, but it was slightly slower in the unilateral case, contradicting Hypothesis H2. The robot achieved significantly fewer collisions per transfer on average than the human operators, confirming Hypothesis H3. The robot traveled a greater distance than the surgeon on average during trials, but less distance than the volunteers, which partially contradicts Hypothesis H4, which predicted that the robot would travel the lowest distance overall.

We are continuing to investigate methods for increasing transfer speeds with additional improvements to sensing, motion planning, and pipelining.

## Acknowledgment

This research was performed at the AUTOLAB at UC Berkeley in affiliation with the Berkeley AI Research (BAIR) Lab, Berkeley Deep Drive (BDD), the Real-Time Intelligent Secure Execution (RISE) Lab, the CITRIS "People and Robots" (CPAR) Initiative, and with UC Berkeley's Center for Automation and Learning for Medical Robotics (Cal-MR).

The authors wish to acknowledge the support of the Technology & Advanced Telemedicine Research Center (TATRC) for their support of this research, performed under contract No. W81XWH-19-C-0096. The research is a component of the Telerobotic Operative Network (TRON) project, being led by SRI International.

The authors are supported in part by donations from Siemens, Google, Toyota Research Institute. The d a Vinci Research Kit was supported by the National Science Foundation, via the National Robotics Initiative (NRI), as part of the collaborative research project "Software Framework for Research in Semi-Autonomous Teleoperation" between The Johns Hopkins University (IIS 1637789), Worcester Polytechnic Institute (IIS 1637759), and the University of Washington (IIS 1637444). Daniel Seita is supported by a Graduate Fellowships for STEM Diversity. We thank Adi Ganapathi, Arnav Gulati, Ashwin Balakrishna, Daniel Brown, Donggun Lee, Ellen Novoseller, Jennifer Grannen, Raven Huang, Zaynah Javed for volunteering.